\documentclass{article}
\usepackage[utf8]{inputenc} 
\usepackage{hyperref}       
\usepackage{url}            
\usepackage{booktabs}       
\usepackage{amsfonts}       
\usepackage{nicefrac}       
\usepackage{microtype}      
\usepackage{amsmath}
\usepackage{graphicx, color}
\usepackage{epstopdf}

\usepackage{bm}
\newtheorem{dfn}{Definition}
\newtheorem{thm}{Theorem}
\newtheorem{lmm}{Lemma}

\newcommand{\argmin}{\mathop{\mathrm{arg~min}}\limits}

\title{Expectation Propagation for t-Exponential Family Using Q-Algebra}

\author{
  Futoshi Futami\\
  The University of Tokyo, RIKEN\\
  \texttt{futami@ms.k.u-tokyo.ac.jp } \\
   \and
   Issei Sato \\
   The University of Tokyo, RIKEN\\
   \texttt{sato@k.u-tokyo.ac.jp} \\
   \and
  Masashi Sugiyama \\
  RIKEN, The University of Tokyo\\
   \texttt{sugi@k.u-tokyo.ac.jp} \\
}
\date{}
\begin{document}
\maketitle
\begin{abstract}
Exponential family distributions are highly useful in machine learning
since their calculation can be performed efficiently through natural parameters.
The exponential family has recently been extended to the \emph{t-exponential family},
which contains Student-t distributions as family members
and thus allows us to handle noisy data well.
However, since the t-exponential family is defined by the \emph{deformed exponential}, we cannot derive an efficient learning algorithm for the t-exponential family such as expectation propagation (EP).
In this paper, we borrow the mathematical tools of \emph{q-algebra} from statistical physics
and show that the \emph{pseudo additivity} of distributions allows us to
perform calculation of t-exponential family distributions through natural parameters. We then develop an \emph{expectation propagation} (EP) algorithm for the t-exponential family,
which provides a deterministic approximation to the posterior or predictive distribution
with simple moment matching.
We finally apply the proposed EP algorithm to
the \emph{Bayes point machine} and \emph{Student-t process classification},
and demonstrate their performance numerically.
\end{abstract}

\section{Introduction}
Exponential family distributions play an important role in machine learning,
due to the fact that their calculation can be performed efficiently and analytically
through natural parameters or expected sufficient statistics [1].
This property is particularly useful in the Bayesian framework
since a conjugate prior always exists for an exponential family likelihood
and the prior and posterior are often in the same exponential family.
Moreover, parameters of the posterior distribution can be evaluated only through natural parameters.

As exponential family members, Gaussian distributions are most commonly used
because their moments, conditional distribution, and joint distribution can be computed analytically.
Gaussian processes are a typical Bayesian method based on Gaussian distributions,
which are used for various purposes such as regression, classification, and optimization [2].
However, Gaussian distributions are sensitive to outliers. It is also known that 
heavier-tailed distributions are often more preferred in practice and 
Student-t distributions would be good alternatives to Gaussian distributions [3]. Similarly Student-t processes would also be promising alternatives to Gaussian processes [4].

The problem of the Student-t distribution is that it does not belong to the exponential family
unlike the Gaussian distribution and thus cannot enjoy good properties of the exponential family.
For this problem,
the exponential family was recently generalized to the \emph{t-exponential family} [5],
which contains Student-t distributions as family members.
Following this line, the Kullback-Leibler divergence was generalized to the \emph{t-divergence},
and approximation methods based on t-divergence minimization have been explored [6].
However, the t-exponential family does not allow us to employ standard useful mathematical tricks,
e.g., logarithmic transformation does not reduce the product of t-exponential family functions into summation.
For this reason, the t-exponential family unfortunately does not inherit an important property of
the original exponential family, that is, calculation can be performed through natural parameters.
Furthermore, while the dimensionality of sufficient statistics is the same as
that of the natural parameters in the exponential family
and thus there is no need to increase the parameter size to incorporate new information [7],
this useful property does not hold in the t-exponential family.

The purpose of this paper is to further explore mathematical properties of natural parameters
of the t-exponential family through \emph{pseudo additivity} of distributions
based on \emph{q-algebra} used in statistical physics [8][9].
More specifically, our contributions in this paper are three-fold:

1. We show that, in the same way as ordinary exponential family distributions,
q-algebra allows us to handle the calculation of t-exponential family distributions
through natural parameters.

2. Our q-algebra based method enables us to extend \emph{assumed density filtering} (ADF) [6]
and develop an algorithm of \emph{expectation propagation} (EP) [10] for the t-exponential family.
In the same way as the original EP algorithm for ordinary exponential family distributions,
our EP algorithm provides a deterministic approximation to the posterior or predictive distribution
for t-exponential family distributions with simple moment matching.

3. We apply the proposed EP algorithm to the \emph{Bayes point machine} [10]
and \emph{Student-t process classification}, and we demonstrate their usefulness
as alternatives to the Gaussian approaches numerically.

\section{t-exponential Family}
In this section, we review the \emph{t-exponential family} [5][6],
which is a generalization of the exponential family.

The t-exponential family is defined as follows,
\begin{eqnarray}
p(x;\theta)&=&\exp_t(\langle\Phi(x),\theta\rangle-g_t(\theta)),
\end{eqnarray}
where $\exp_t(x)$ the \emph{deformed exponential function} defined as
\begin{eqnarray}
\exp_t(x)&=& \left\{
\begin{array}{ll}
\exp(x)&\mathrm{if~} t=1,\\
\lbrack1+(1-t)x\rbrack ^{\frac{1}{1-t}}&\mathrm{otherwise},
\end{array}\right.
\end{eqnarray}
and $g_t (\theta)$ is the log-partition function that satisfies
\begin{equation}
\nabla_\theta g_t(\theta)=\mathbb{E}_q \lbrack \Phi(x) \rbrack.
\end{equation}
The notation $\mathbb{E}_q$ denotes the expectation over $q(x)$,
where $q(x)$ is the \emph{escort distribution} of $p(x)$ defined as
\begin{equation}
q(x)=\frac{p(x)^t}{\int p(x)^t \mathrm{d}x}.
\end{equation}
We call $\theta$ a \emph{natural parameter} and $\Phi(x)$ a \emph{sufficient statistics}.

Let us express the $k$-dimensional Student-t distribution with $v$ degrees of freedom as
\begin{eqnarray}
\mathrm{St}(x; v, \mu, \Sigma)=\frac{\Gamma ((v+k)/2)}{(\pi v)^{k/2}\Gamma (v/2)|\Sigma|^{1/2}}\biggl(1+(x-\mu)^{\top}|v\Sigma|^{-1}(x-\mu) \biggr)^{-\frac{v+k}{2}},
\end{eqnarray}
where $\Gamma (x)$ is the gamma function, $|A|$ is the determinant of matrix $A$,
and $A^\top$ is the transpose of matrix $A$.
We can confirm that the Student-t distribution is a member of the t-exponential family
as follows. First, we have
\begin{eqnarray}
\mathrm{St}(x; v, \mu, \Sigma)=\bigl(\Psi+\Psi\cdot (x-\mu)^{\top}(v\Sigma)^{-1}(x-\mu)\bigr)^{\frac{1}{1-t}},\\
{\rm where\ \ }\Psi=\Biggl(\frac{\Gamma((v+k)/2)}{(\pi v)^{k/2}\Gamma(v/2)|\Sigma|^{1/2}}\Biggr)^{1-t}.\label{eq:eq10}
\end{eqnarray}
Note that relation $-(v+k)/2=1/(1-t)$ holds, from which we have
\begin{eqnarray}
\langle\Phi(x),\theta\rangle&=&\Biggl(\frac{\Psi}{1-t}\Biggr)(x^\top Kx-2\mu^\top Kx),\\
g_t(\theta)&=&-\Biggl(\frac{\Psi}{1-t}\Biggr)(\mu^\top K\mu+1)+\frac{1}{1-t},
\end{eqnarray}
where $K=(v\Sigma)^{-1}$. Second, we can express the Student-t distribution
as a member of the t-exponential family:
\begin{eqnarray}
\mathrm{St}(x; v, \mu, \Sigma)=\bigl(1+(1-t)\langle\Phi(x),\theta\rangle-g_t(\theta)\bigr)^{\frac{1}{1-t}}=\exp_t\bigl(\langle\Phi(x),\theta\rangle-g_t(\theta)\bigr).
\end{eqnarray}
If $t=1$, the deformed exponential function is reduced to the ordinary exponential,
and therefore the t-exponential family is reduced to the ordinary exponential family,
which corresponds to the Student-t distribution with infinite degrees of freedom. 
For t-exponential family distributions, a divergence is defined as follows [6]:
\begin{eqnarray}
D_t(p \| \tilde{p} )=\int q(x)\ln _t p(x)-q(x)\ln _t \tilde{p} (x)\mathrm{d}x,
\end{eqnarray}
where $\ln_tx:=\frac{x^{1-t}-1}{1-t} \ \ (x\geq0,t \in \mathbb{R}^+)$. 
This is called the \emph{t-divergence} and $q(x)$ is the escort function of $p(x)$.

\section{Assumed Density Filtering and Expectation Propagation}
We briefly review the assumed density filtering (ADF) and expectation propagation (EP) [10]. 

Let $D=\{(x_1,y_1),\ldots,(x_{i},y_{i})\}$ be input-output paired data.
We denote the likelihood for the $i$-th data as $l_i(w)$
and the prior distribution of parameter $w$ as $p^{(0)}(w)$.
The total likelihood is given as $\prod_i l_i(w)$
and the posterior distribution can be expressed as $p(w|D)\propto p^{(0)}(w)\prod_i l_i(w)$.

\subsection{Assumed Density Filtering}
ADF is an online approximation method for the posterior distribution.

Suppose that $i-1$ samples $(x_1,y_1),\ldots,(x_{i-1},y_{i-1})$
have already been processed
and an approximation to the posterior distribution, $\tilde{p}_{i-1}(w)$, has already been obtained.
Given the $i$-th sample $(x_i,y_i)$,
the posterior distribution $p_i(w)$ can be obtained as
\begin{eqnarray}
p_i(w)\propto \tilde{p}_{i-1}(w)l_i(w).
\end{eqnarray}
Since the true posterior distribution $p_i(w)$ cannot be obtained analytically,
it is approximated in ADF by minimizing the Kullback-Leibler (KL) divergence
from $p_i(w)$ to its approximation:
\begin{eqnarray}
\tilde{p}_i=\argmin _{\tilde{p}} \mathrm{KL}(p_i \| \tilde{p} ).
\end{eqnarray}
Note that if $\tilde{p}$ is an exponential family members,
the above calculation is reduced to moment matching.

\subsection{Expectation Propagation}
Although ADF is an effective method for online learning,
it is not favorable for non-online situations,
because the approximation quality depends heavily on the permutation of data.
To overcome this problem, EP was proposed. 

In EP, an approximation of the posterior that contains whole data terms is prepared beforehand,
typically as a product of data-corresponding terms:
\begin{eqnarray}
\tilde{p}(w) = \frac{1}{Z}p^{(0)}(w)\prod_i \tilde{l}_i(w),
\end{eqnarray}
where $Z$ is the normalizing constant.
In the above expression, factor $\tilde{l}_i(w)$,
which is often called a \emph{site approximation},
corresponds to the local likelihood $l_i(w)$.
If $\tilde{l}_i(w)$ is an exponential family member,
the total approximation also belongs to the exponential family.

Differently from ADF, in EP, these site approximations are updated iteratively in four steps as follows.
First, when we update site $\tilde{l}_j(w)$,
we eliminate the effect of site $j$ from the total approximation as
\begin{eqnarray}
\tilde{p}^{\backslash j}(w) \propto \frac{\tilde{p}(w)}{\tilde{l}_j(w)},
\end{eqnarray}
where $\tilde{p}^{\backslash j}(w)$ is often called a \emph{cavity distribution}.
If an exponential family distribution is used,
the above calculation is reduced to the subtraction of natural parameters. 
Second, we incorporate likelihood $l_j(w)$
by minimizing the divergence $\mathrm{KL}$$(\tilde{p}^{\backslash j}(w)$$l_j(w)/Z^{\setminus j}$$\| \tilde{p}(w) )$,
where $Z^{\setminus j}$ is the normalizing constant.
Note that this minimization is reduced to moment matching for the exponential family. After this step, we obtain $\tilde{p}(w)$. Third, we exclude the effect of terms other than $j$,
which is equivalent to calculating a cavity distribution as $\tilde{l}_j(w)^{\mathrm{new}} \propto \frac{\tilde{p}(w)}{\tilde{p}^{\setminus j}(w)}$
Finally, we update the site approximation by replacing  $\tilde{l}_j(w)$ by $\tilde{l}_j(w)^{\mathrm{new}}$.

Note again that calculation of EP is reduced to addition or subtraction of
natural parameters for the exponential family.

\subsection{ADF for t-exponential Family}
ADF for the t-exponential family was proposed in [6],
which uses the \emph{t-divergence} instead of the KL divergence:
\begin{eqnarray}
\tilde{p}=\argmin _{p'} D_t(p \| p' )=\int q(x)\ln _t p(x)-q(x)\ln _t p' (x;\theta)\mathrm{d}x.
\end{eqnarray}
When an approximate distribution is chosen from the t-exponential family,
we can utilize the property $\nabla_\theta g_t(\theta)=\mathbb{E}_{\tilde{q}} (\Phi(x))$,
where $\tilde{q}$ is the escort function of $q(x)$.
Then, minimization of the t-divergence yields
\begin{eqnarray}
\mathbb{E}_q \lbrack \Phi(x) \rbrack=\mathbb{E}_{\tilde{q}} \lbrack \Phi(x) \rbrack.
\end{eqnarray}
This is moment matching, which is a celebrated property of the exponential family. 
Since the expectation is respect to the escort function,
this is called \emph{escort moment matching}. 

As an example, let us consider the situation where
the prior is the Student-t distribution
and the posterior is approximated by the Student-t distribution: 
$p(w|D)\cong \tilde{p}(w)=\mathrm{St}(w; \tilde{\mu}, \tilde{\Sigma},v)$.
The approximated posterior $\tilde{p}_{i}(w)=\mathrm{St}(w; \tilde{\mu}^{(i)}, \tilde{\Sigma}^{i},v)$
can be obtained by minimizing the t-divergence
from $p_i(w) \propto \tilde{p}_{i-1}(w)\tilde{l}_i(w)$ as
\begin{eqnarray}
\argmin _{\mu',\Sigma'} D_t(p_i (w) \| \mathrm{St}(w; \mu', \Sigma',v) ).
\end{eqnarray}
This allows us to obtain an analytical update expression
for t-exponential family distributions.  

\section{Expectation Propagation for t-exponential Family}
As shown in the previous section, ADF has been extended 
to EP (which resulted in moment matching for the exponential family)
and to the t-exponential family 
(which yielded escort moment matching for the t-exponential family).
In this section, we combine these two extensions and propose EP for the t-exponential family.

\subsection{Pseudo Additivity and Q-Algebra}
Differently from ordinary exponential functions,
\emph{deformed} exponential functions do not satisfy the product rule:
\begin{eqnarray}
\exp_t(x)\exp_t(y) \neq \exp_t(x+y).
\end{eqnarray}
For this reason, the cavity distribution cannot be computed analytically
for the t-exponential family.

On the other hand, the following equality holds for the deformed exponential functions:
\begin{eqnarray}
\exp_t(x)\exp_t(y) = \exp_t(x+y+(1-t)xy),
\end{eqnarray}
which is called \textit{pseudo additivity}.

In statistical physics [8][9], a special algebra called \textit{q-algebra} has been developed to
handle a system with pseudo additivity.
We will use the q-algebra for efficiently handling t-exponential distributions.

\begin{dfn}[q-product]
Operation $\otimes _q$ called the \emph{q-product} is defined as
\begin{equation}
x\otimes _q y:= \left\{
\begin{array}{ll}
[x^{1-q}+y^{1-q}-1]^\frac{1}{1-q}&\mathrm{if~} x>0, y>0, x^{1-q}+y^{1-q}-1>0,\\
0&\mathrm{otherwise}.
\end{array}\right.
\end{equation}
\end{dfn}
\begin{dfn}[q-division]
Operation $\oslash _q$  called the \emph{q-division} is defined as
\begin{equation}
x\oslash _q y:= \left\{
\begin{array}{ll}
[x^{1-q}-y^{1-q}-1]^\frac{1}{1-q}&\mathrm{if~}\ x>0, y>0, x^{1-q}-y^{1-q}-1>0,\\
0&\mathrm{otherwise}.
\end{array}\right.
\end{equation}
\end{dfn}
\begin{dfn}[q-logarithm]
The \emph{q-logarithm} is defined as
\begin{equation}
\ln_qx:=\frac{x^{1-q}-1}{1-q} \ \ (x\geq0,q \in \mathbb{R}^+).
\end{equation}
\end{dfn}

The q-division is the inverse of the q-product (and visa versa),
and the deformed q-logarithm is the inverse of the deformed exponential (and visa versa).
From the above definitions, the deformed logarithm and exponential satisfy
the following relations:
\begin{eqnarray}
\ln_q(x \otimes _q y)=\ln_q x+\ln_q y,\\
\exp _q (x) \otimes _q \exp _q (y)=\exp _q (x+y),
\end{eqnarray}
which are called the \emph{q-product rules}.
Also for the q-division, similar properties hold:
\begin{eqnarray}
\ln_q(x \oslash _q y)=\ln_q x-\ln_q y,\\
\exp _q (x) \oslash _q \exp _q (y)=\exp _q (x-y),
\end{eqnarray}
which are called the \emph{q-division rules}.

\subsection{EP for t-exponential Family}
The q-algebra allows us to recover many useful properties from the ordinary exponential family.
For example, the q-product of t-exponential family distributions
yields an unnormalized t-exponential distribution:
\begin{equation}
\exp_t(\langle\Phi(x),\theta_1\rangle-g_t(\theta_1))\otimes_t \exp_t(\langle\Phi(x),\theta_2\rangle-g_t(\theta_2))=\exp_t(\langle\Phi(x),(\theta_1+\theta_2)\rangle-\tilde{g}_t(\theta_1,\theta_2)).
\end{equation}

Based on this q-product rule, we develop EP for the t-exponential family.
Consider the situation where prior distribution $p^{(0)}(w)$ is a member of the t-exponential family.
As an approximation to the posterior, we choose a t-exponential family distribution $\tilde{p}(w;\theta)=\exp_t(\langle\Phi(w),\theta\rangle-g_t(\theta))$.
In the original EP for the ordinary exponential family, we considered
an approximate posterior of the form $\tilde{p}(w) \propto p^{(0)}(w)\prod_i \tilde{l}_i(w)$, that is, we factorized the posterior to 
a product of site approximations corresponding to data.
On the other hand, in the case of the t-exponential family,
we propose to use the following form called the \emph{t-factorization}: 
\begin{eqnarray}
\tilde{p}(w) \propto p^{(0)}(w) \otimes _t \prod_i \otimes _t \tilde{l}_i(w).
\end{eqnarray}
where $\tilde{l}_i(w)$ is an unnormalized t-exponential family $\tilde{C}_i\otimes \exp_t(\langle\Phi(w),\theta\rangle)$.
The t-factorization is reduced to the original factorization form when $t=1$.

This t-factorization enables us to calculate EP update rules through natural parameters
for the t-exponential family in the same way as the ordinary exponential family.
More specifically, consider the case where factor $j$ of the t-factorization
is updated following four steps as in the same way of ordinal EP.
(I) First, we calculate the cavity distribution by using the q-division as
\begin{eqnarray}
\tilde{p}^{\backslash j}(w) \propto \tilde{p}(w)\oslash _t\tilde{l}_j(w)\propto p^{(0)}(w) \otimes _t \prod_{i\neq j} \otimes _t \tilde{l}_i(w).
\end{eqnarray}
The above calculation is reduced to subtraction of natural parameters
by using the q-algebra rules:
\begin{eqnarray}
\theta ^{\setminus j}=\theta-\theta^{(j)}.
\end{eqnarray}
(II) Second step is the inclusion of site likelihood $l_j(w)$, which can be performed by $\tilde{p}^{\backslash j}(w) l_j(w)$.
The site likelihood $l_j(w)$ is incorporated to approximate the posterior
by the ordinary product not the q-product. Thus moment matching is performed to obtain a new approximation.
For this purpose, the following theorem is useful.
\begin{thm}\label{thm:eta}
The expected sufficient statistics
where $\eta=\nabla_{\theta}g_t(\theta)=\mathbb{E}_{\tilde{q}} \lbrack \Phi(w) \rbrack$, 
can be derived as
\begin{eqnarray}
&&\eta=\eta^{\setminus j}+\frac{1}{Z_2}\nabla_{\theta^{\setminus j}}Z_1,\\
&&\mathrm{where}\ \ \ Z_1=\int \tilde{p}^{\backslash j}(w) (l_j(w))^t \mathrm{d}w,\ \ \ \ Z_2=\int \tilde{q}^{\backslash j}(w) (l_j(w))^t \mathrm{d}w.
\end{eqnarray}
\end{thm}
A proof of Theorem~\ref{thm:eta} is given in Appendix~\ref{proof:thm:eta} of supplemental material. After moment matching, we obtain the approximation,  $\tilde{p}_{\mathrm{new}}(w)$.\\
(III) Third, we exclude the effect of sites other than $j$. This is achieved by $\tilde{l}_j ^{\mathrm{new}}(w) \propto \tilde{p}_{\mathrm{new}}(w)\oslash _t \tilde{p}^{\backslash j}(w)$,
which is reduced to subtraction of natural parameter $\theta_{\mathrm{new}}^{\setminus j}=\theta^{\mathrm{new}}-\theta^{\setminus j}$.\\
(IV)Finally, we update the site approximation by replacing $\tilde{l}_j(w)$ by $\tilde{l}_j(w)^\mathrm{new}$.

These four steps are our proposed EP method for the t-exponential family.
As we have seen, these steps are reduced to the ordinary EP steps if $t=1$.
Thus, the proposed method can be regarded as
an extention of the original EP to the t-exponential family.

\subsection{Marginal Likelihood for t-exponential Family}
In the above, we omitted the normalization term of the site approximation
to simplify the derivation.
Here, we derive the marginal likelihood,
which requires us to explicitly take into account the normalization term $\tilde{C}_i$:
\begin{eqnarray}
\tilde{l}_i(w)=\tilde{C}_i\otimes \exp_t(\langle\Phi(w),\theta\rangle).
\end{eqnarray}
We assume that this normalizer corresponds to $Z_1$,
which is the same assumption as that for the ordinary EP.
To calculate $Z_1$, we use the following theorem
(its proof is available in Appendix~\ref{proof-thm:Student-t} of supplemental material):
\begin{thm}\label{thm:Student-t}
For the Student-t distribution, we have
\begin{eqnarray}
\int \exp_t(\langle\Phi(w),\theta\rangle-g)\mathrm{d}w
=\Bigl(\exp_t(g_t(\theta)/\Psi-g/\Psi)\Bigr)^{\frac{3-t}{2}},
\end{eqnarray}
where $g$ is a constant, $g(\theta)$ is the log partition function and $\Psi$ is defined in \eqref{eq:eq10}.
\end{thm}
This theorem yields
\begin{eqnarray}
\log _t Z_1^{\frac{2}{3-t}}=g_t(\theta)/\Psi-g_t^{\setminus j}(\theta)/\Psi+\log_t \tilde{C}_j/\Psi,
\end{eqnarray}
and therefore the marginal likelihood can be calculated as follows (see Appendix~\ref{sec:marginal-likelihood} for details):
\begin{eqnarray}
Z_{\mathrm{EP}}=\int p^{(0)}(w) \otimes _t \prod_i \otimes _t \tilde{l}_i(w)dw
=\Biggl(\exp_t\Bigl(\sum _i \log _t\tilde{C}_i/\Psi+g_t(\theta)/\Psi-g_t^{\mathrm{prior}}(\theta)/\Psi\Bigr)\Biggr)^{\frac{3-t}{2}}.
\end{eqnarray}
By substituting $\tilde{C}_i$, we obtain the marginal likelihood.
Note that, if $t=1$, the above expression of $Z_{\mathrm{EP}}$ is reduced to
the ordinary likelihood expression in [7]. 
Therefore, this likelihood can be regarded as a generalization of
the ordinary exponential family likelihood to the t-exponential family.

In Appendices~\ref{sec:BPM} and \ref{sec:t-process} of supplemental material,
we derive specific EP algorithms for
the \emph{Bayes point machine} (BPM) [6] and \emph{Student-t process classification}.

\section{Numerical Illustration}
In this section, we numerically illustrate the behavior
of our proposed EP applied to BPM and Student-t process classification.
Suppose that data  $(x_1,y_1),\dots,(x_{n},y_{n})$ are given, where $y_i\in\{+1,-1\}$ expresses
a class label for sample $x_i$.
We consider a model whose likelihood term can be expressed as
\begin{eqnarray}
l_i (w) = p(y_i | x_i,w)=\epsilon + (1-2\epsilon)\Theta(y_i\langle w,x_i\rangle),
\end{eqnarray}
where $\Theta(x)$ is the step function taking $1$ if $x>0$ and $0$ otherwise.

\subsection{BPM}
We compare EP and ADF to confirm that EP does not depend on data permutation.
We generate a toy dataset in the following way:
1000 data points $x$ are generated from Gaussian mixture model $0.05N(x;[1,1],0.05I)+0.25N(x;[-1,1],0.05I)+0.45N(x;[-1,-1],0.05I)+0.25N(x;[1,-1],0.05I)$,
where $N(x;\mu,\Sigma)$ denotes the Gaussian density with respect to $x$
with mean $\mu$ and covariance matrix $\Sigma$, and $I$ is the identity matrix.
For $x$, we assign label $y=+1$ when $x$ comes from $N(x;[1,1],0.05I)$ or $N(x;[1,-1],0.05I)$
and label $y=-1$ when $x$ comes from $N(x;[-1,1],0.05I)$ or $N(x;[-1,-1],0.05I)$. We evaluate the dependence of the performance of BPM (see Appendix~\ref{sec:BPM} of supplemental material for details)
on the data permutation. 

\begin{figure}[t]
  \centering
    \begin{tabular}{@{}c@{\ }c@{\ }c@{}}
          \includegraphics[clip, width=0.4\textwidth]{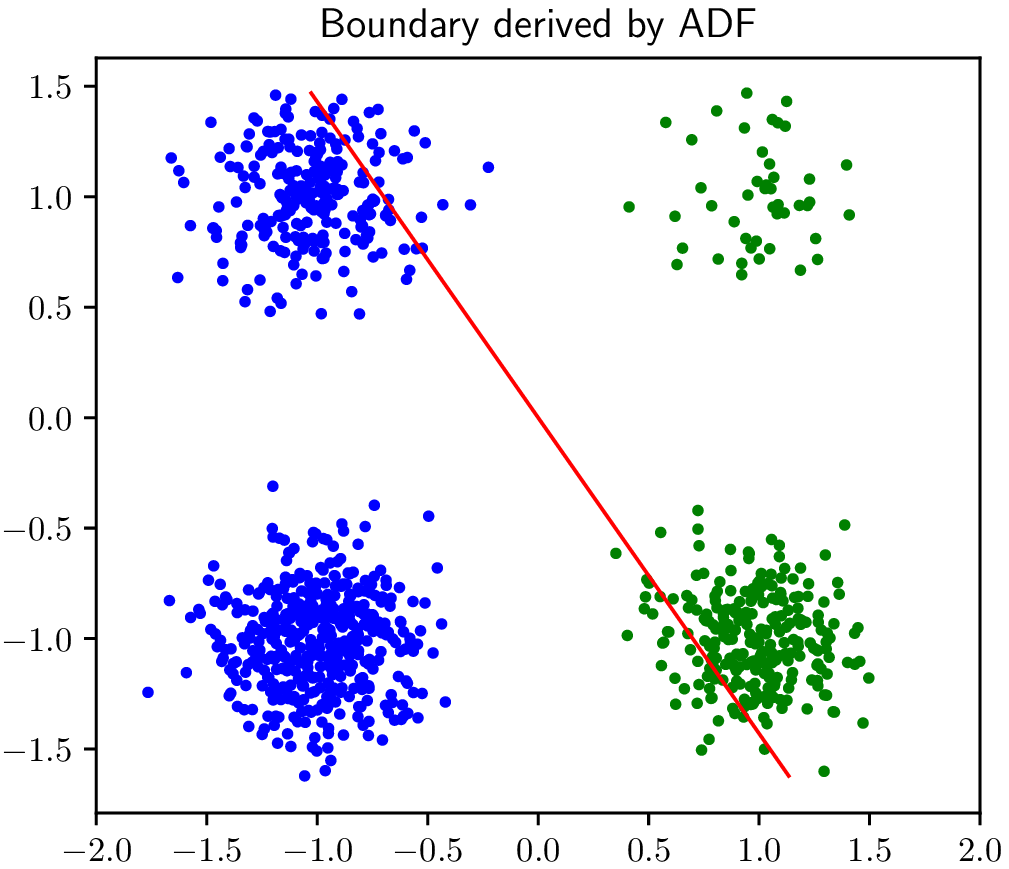}
          \includegraphics[clip, width=0.4\textwidth]{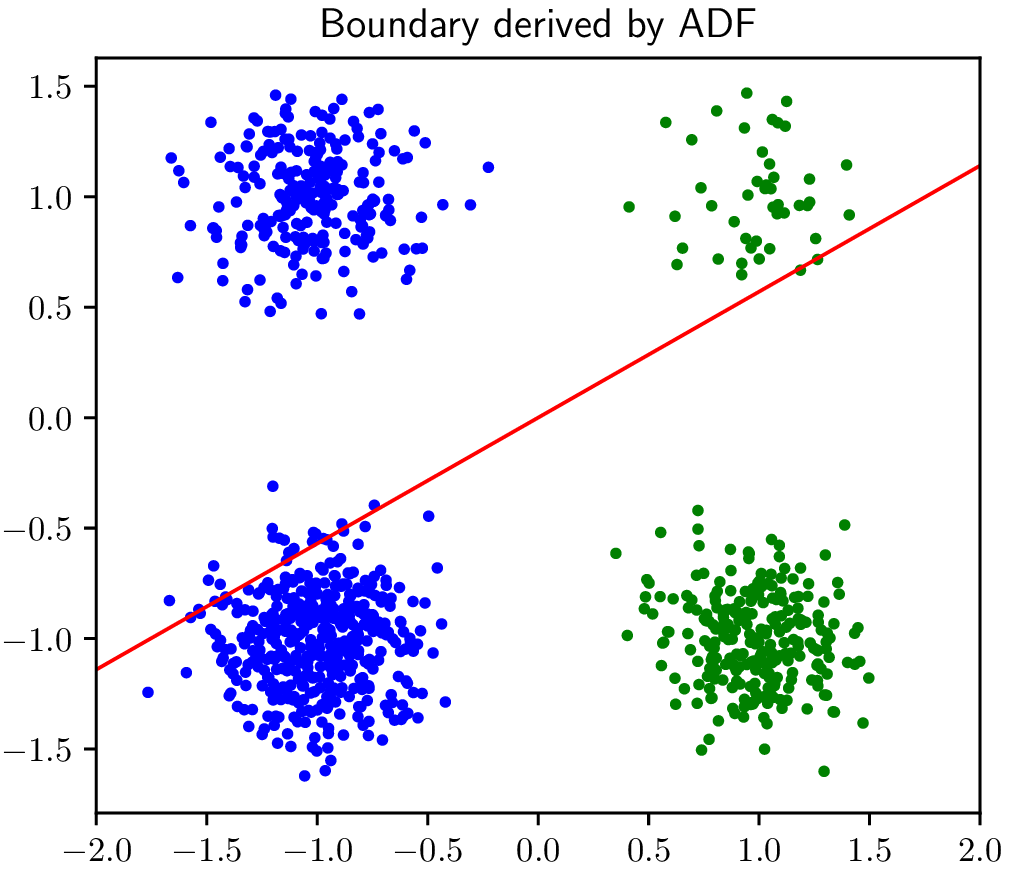}\\
          \includegraphics[clip, width=0.4\textwidth]{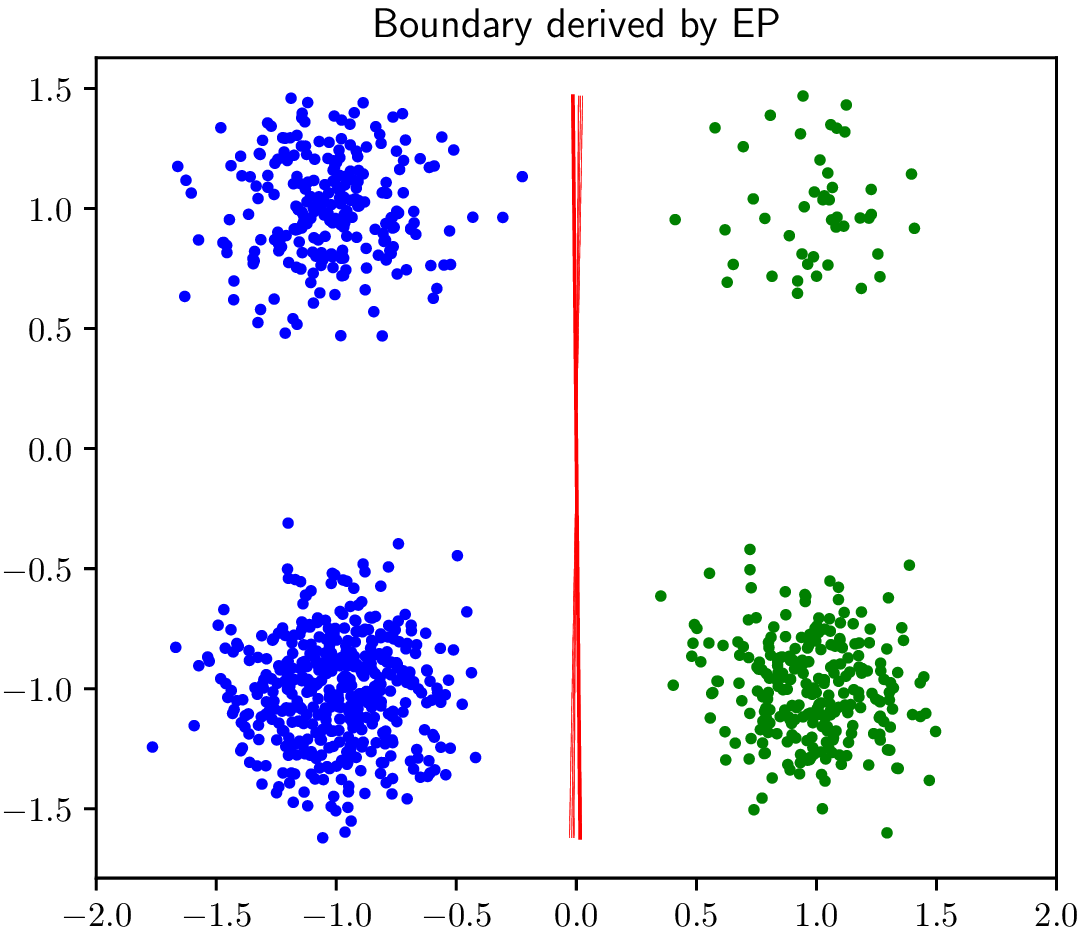}
    \end{tabular}
    \caption{Classification boundaries derived by ADF and EP.}
    \label{fig:Posterir-distribution-BPM}
\end{figure}

Fig.\ref{fig:Posterir-distribution-BPM} shows labeled samples by blue and green points, decision
boundaries by red lines which are derived from ADF and EP for the
Student-t distribution with $v=10$ by changing data permutations. The top two graph shows obvious dependence on data permutation by ADF (to clarify the dependence on data permutation, we showed the most different boundary in the figure), while the bottom graph exhibits almost no dependence on data permutations by EP.

\subsection{Student-t Process Classification}

We compare the robustness of Student-t process classification
and Gaussian process classification.

We apply our EP method to Student-t process binary classification,
where the latent function follows the Student-t process
(see Appendix~\ref{sec:t-process} of supplemental material for details).
We compare this model with Gaussian process binary classification with the same likelihood term.
Since the posterior distribution cannot be obtained analytically
even for the Gaussian process,
we use EP for the ordinary exponential family to approximate the posterior.

We use a two-dimensional toy dataset, 
where we generate a two-dimensional data point $x_i$ ($i=1,\ldots,200$)
following the normal distribution with mean $[y_i,y_i]$ and unit variance,
where $y_i\in\{+1,-1\}$ is the class label for $x_i$ determined randomly.
We add three outliers to the dataset and evaluate the robustness against outliers.
In the experiment, we used the Gaussian kernel, and we used $v=10$ for Student-t processes.

\begin{figure}[t]
  \centering
    \begin{tabular}{@{}c@{\ }c@{}}
      \includegraphics[clip, width=0.8\textwidth]{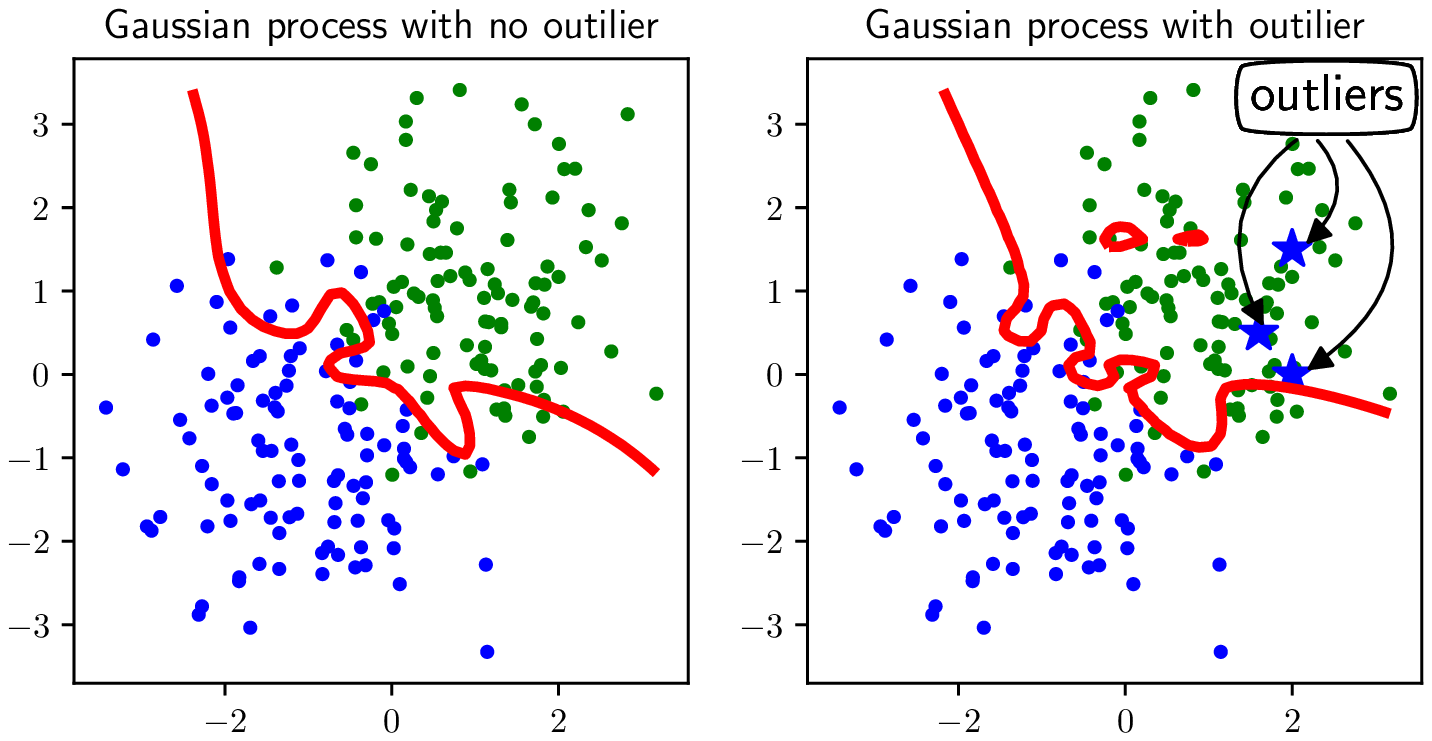}\\
      \includegraphics[clip, width=0.8\textwidth]{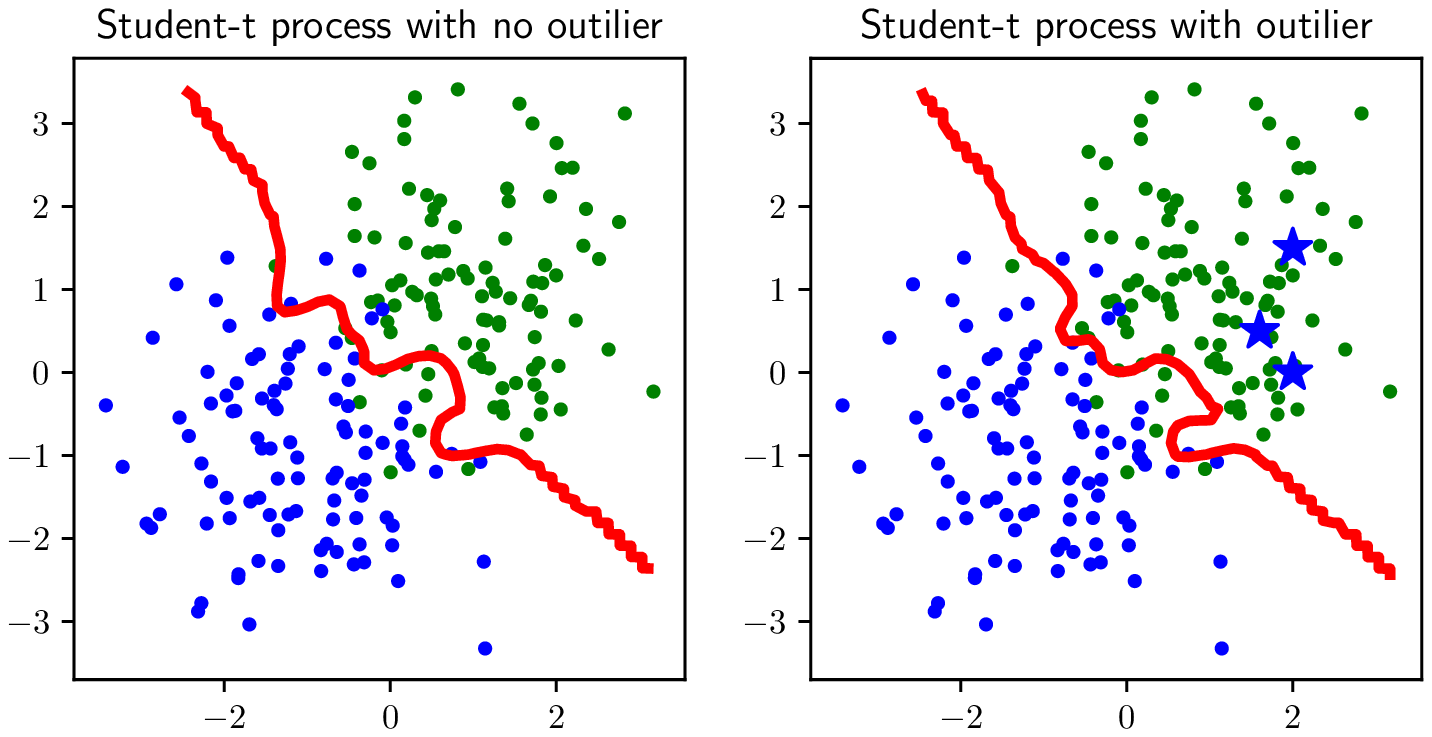}
    \end{tabular}
    \caption{Classification boundaries.}
    \label{fig:Stc}
\end{figure}

Fig.\ref{fig:Stc} shows the labeled samples by blue and green points,
the obtained decision boundaries by red lines, and added outliers by blue stars.
As we can see, the decision boundaries obtained by the Gaussian process classifier
is heavily affected by outliers, while those obtained by the Student-t process classifier
are more stable.
Thus, as expected, Student-t process classification is more robust
against outliers compared to Gaussian process classification,
thanks to the heavy-tailed structure of the Student-t distribution.

\section{Conclusions}
In this work, we enabled the \emph{t-exponential family} to inherit the important property of the exponential family that calculation can be efficiently performed thorough natural parameters by using the \emph{q-algebra}. By using this natural parameter based calculation, we developed EP for the t-exponential family by introducing the \emph{t-factorization} approach. The key concept of our proposed approach is that the t-exponential family have \textit{pseudo additivity}.
When $t=1$, our proposed EP for the t-exponential family is reduced to
the original EP for the ordinary exponential family and
 t-factorization yields ordinary data-dependent factorization.
Therefore, our proposed EP method can be viewed as a generalization of the original EP.
Through illustrative experiments, we confirmed that our proposed EP applied to the Bayes point machine
can overcome the drawback of ADF, i.e., the proposed EP method is independent of data permutations.
We also experimentally illustrated that our proposed EP applied to Student-t process classification
exhibited high robustness to outliers compared to Gaussian process classification.

In our future work, we will further extend the proposed EP method 
to more general message passing methods or double-loop EP.
We would like also to make our method more scalable to large datasets
and develop another approximating method such as variational inference.

\newpage
\appendix
\section{Proof of Theorem~\ref{thm:eta}}\label{proof:thm:eta}
\begin{eqnarray}
\nabla_{\theta^{\setminus j}}Z_1&=&\nabla_{\theta^{\setminus j}}\int \tilde{p}^{\backslash j}(w) l_j(w)^t dw \nonumber \\
&=&\int (\phi(w)-\nabla_{\theta^{\setminus j}}g_t(\theta ^{\setminus j}))\tilde{q}^{\backslash j}(w) l_j(w)^t dw \nonumber \\
&=&\int \phi(w)\tilde{q}^{\backslash j}(w) l_j(w)^t dw - \nabla_{\theta^{\setminus j}}g_t(\theta ^{\setminus j})\int \tilde{q}^{\backslash j}(w) l_j(w)^t dw\nonumber 
\end{eqnarray}
Using the definition $Z_2=\int \tilde{q}^{\backslash j}(w) (l_j(w))^t dw$, and $\eta=\nabla_{\theta}g_t(\theta)$,
\begin{eqnarray}
\nabla_{\theta^{\setminus j}}Z_1&=&\eta Z_2-\eta^{\setminus j}Z_2 \nonumber
\end{eqnarray}
Therefore,
\begin{eqnarray}
\eta=\eta^{\setminus j}+\frac{1}{Z_2}\nabla_{\theta^{\setminus j}}Z_1\nonumber.
\end{eqnarray}

\section{Proof of Theorem~\ref{thm:Student-t}}\label{proof-thm:Student-t}
Here, we consider a one-dimensional case, but we can consider this in the same way as for a multivariate case. Considering the unnormalized t-exponential family, $\exp_t(\langle\Phi(w),\theta\rangle-g)$, and $g$ is a constant, not a true log partition function. We integrate this expression as follows,
\begin{eqnarray}
\int^{\infty} _{-\infty} \exp_t(\langle\Phi(w),\theta\rangle-g)dw&=&\int^{\infty} _{-\infty} (1+\Psi(-2\mu^\top K w+w^\top Kw)-(1-t)g)^{\frac{1}{1-t}}dw \nonumber \\
&=&\int^{\infty} _{-\infty} (1-\Psi \mu^\top K\mu-(1-t)g+\Psi(w-\mu)^\top K(w-\mu))^{\frac{1}{1-t}}dw \nonumber \\
&=&(1-\Psi \mu^\top K\mu-(1-t)g)^{\frac{1}{1-t}}\int^{\infty} _{-\infty} \Bigl(1+\frac{\Psi(x-\mu)^\top K(x-\mu)}{1-\Psi \mu^\top K\mu-(1-t)g}\Bigr)^{\frac{1}{1-t}}dw \nonumber 
\end{eqnarray}
Here, for simplicity, we put $(1-\Psi \mu^\top K\mu-(1-t)g)=A$, and use the formula, $\int ^{\infty} _0 \frac{x^m}{(1+x^2)^n}dx = \frac{1}{2}B\bigl(\frac{2n-m-1}{2},\frac{m+1}{2}\bigr)$, where $B$ denote the beta function. We can get the expression,
\begin{eqnarray}
\int^{\infty} _{-\infty} \exp_t(\langle\Phi(w),\theta\rangle-g)dw&=&\frac{1}{2}B\Bigl(\frac{3-t}{2(t-1)},\frac{1}{2}\Bigr)\Bigl(\frac{\Psi}{A}K\Bigr)^{-\frac{1}{2}}A^{\frac{1}{1-t}} \nonumber
\end{eqnarray}
We can proceed with the calculation by using the definition of $\Psi$, $B(x,y)=\frac{\Gamma(x)\Gamma(y)}{\Gamma(x+y)}$, and$\Gamma(\frac{1}{2})=\sqrt{\pi}$ as follows,
\begin{eqnarray}
\int^{\infty} _{-\infty} \exp_t(\langle\Phi(w),\theta\rangle-g)dw&=&\Psi^{-\bigl(\frac{1}{2}+\frac{1}{1-t}\bigr)}A^{\frac{1}{2}+\frac{1}{1-t}} \nonumber
\end{eqnarray}
Here, by using the definition of $A$ and the true log partition function $g_t(\theta)=\frac{1}{1-t}\bigl(1-\Psi(\mu^\top K\mu+1)\bigr)$,
\begin{eqnarray}
A^{\frac{1}{2}+\frac{1}{1-t}}&=&(1-\Psi \mu^\top K\mu-(1-t)g)^{\frac{1}{2}+\frac{1}{1-t}} \nonumber \\
&=&(\Psi+(1-t)(g_t(\theta)-g))^{\frac{1}{2}+\frac{1}{1-t}} \nonumber \\
&=&\Psi^{\frac{1}{2}+\frac{1}{1-t}}(1+(1-t)(g_t(\theta)-g)/\Psi)^{\frac{1}{2}+\frac{1}{1-t}} \nonumber
\end{eqnarray}
Therefore, by substituting this expression into the above integral result, we get the following.
\begin{eqnarray}
\int^{\infty} _{-\infty} \exp_t(\langle\Phi(w),\theta\rangle-g)dw=\Bigl(\exp_t(g_t(\theta)/\Psi-g/\Psi)\Bigr)^{\frac{3-t}{2}}\nonumber 
\end{eqnarray}

\section{Deriving the Marginal likelihood}\label{sec:marginal-likelihood}
\begin{eqnarray}
Z_{\rm{EP}}&=&\int p^{(0)}(w) \otimes _t \prod_i \otimes _t \tilde{l}_i(w)dw \nonumber \\
&=&\int \exp_t\Bigl(\sum _i \log _t\tilde{C}_i+\langle\Phi(w),\theta\rangle-g_t^{\rm{prior}}(\theta)\Bigr)dw \nonumber \\
&=&\Biggl(\exp_t\Bigl(\sum _i \log _t\tilde{C}_i/\Psi+g_t(\theta)/\Psi-g_t^{\rm{prior}}(\theta)/\Psi\Bigr)\Biggr)^{\frac{3-t}{2}}.\nonumber 
\end{eqnarray}

\section{Bayes Point Machine}\label{sec:BPM}
In this section, we show the details of the update rule of ADF and EP for the Bayes point machine.
\subsection{ADF update rule for BPM}
The detailed update rules of ADF for BPM in t-exponential family are derived in [6].
\begin{eqnarray}
\mu^i&=&E_q\lbrack w\rbrack=\mu ^{i-1}+\alpha y_i\Sigma ^{i-1} x_i\\
\Sigma^i&=&E_q\lbrack ww^\top\rbrack-E_q\lbrack w\rbrack E_q\lbrack w^\top\rbrack =r\Sigma^{i-1}-(\Sigma^{i-1} x_i)\biggl(\frac{\alpha y_i\langle x_i,\mu^i\rangle}{x_i^{\top} \Sigma^{i-1} x_i}\biggr)(\Sigma^{i-1} x_i)^\top,
\end{eqnarray}
where $\tilde{q}_i(w)\propto\tilde{p}_i(w) ^t$, $q_i(w)\propto\tilde{p}_{i-1}(w) ^t(l_i(w))^t$, and 
\begin{eqnarray}
z&=&\frac{\alpha y_i\langle x_i,\mu^{i-1}\rangle}{\sqrt{x_i^{\top} \Sigma^{i-1} x_i}}\\
Z_1&=&\int \tilde{p}_{i-1}(w)(l_{i}(w))^tdw=\epsilon^t+((1-\epsilon)^t-\epsilon^t)\int_{\infty} ^z \mathrm{St}(z;0,1,v)\\
Z_2&=&\int \tilde{q}_{i-1}(w)(l_i(w))^tdw=\epsilon^t+((1-\epsilon)^t-\epsilon^t)\int_{\infty} ^z \mathrm{St}(z;0,v/(v+2),v+2)\\
r&=&\frac{Z_1}{Z_2}\\
\alpha&=&\frac{((1-\epsilon)^t-\epsilon^t)\mathrm{St}(z;0,1,v)}{Z_2\sqrt{x_i ^\top\Sigma^{i-1}x_i}}
\end{eqnarray}

\subsection{EP update rule for BPM}
As for the EP update rule, natural parameters of Student-t distribution $\mathrm{St}(w; v, \mu, \Sigma)$ is $[\theta_1, \theta_2]$, 
\begin{eqnarray}
\theta_1&=&-2\frac{\Psi K\mu}{1-t}\\
\theta_2&=&\frac{\Psi K}{1-t}
\end{eqnarray}
where, $K=(v\Sigma)^{-1}$. From these, we can calculate EP update rules through $\Psi K\mu$ and $\Psi K$. 

For the BPM, we consider that the whole approximation is $k$-dimensional $\mathrm{St}(w;m_w,V_w,v)$, and the site approximation as one-dimensional Student-t like function, $\exp_t(\langle\Phi(w),\theta\rangle)$, where $\langle\Phi(w),\theta\rangle=\frac{\Psi _i}{1-t}\bigl((w^\top x_i)^\top (v\sigma_i)^{-1}(w^\top x_i)-2m_i (\tilde{v}\sigma_i)^{-1}(w^\top x_i)\bigr)\propto \frac{\Psi _i}{1-t}\tilde{v}^{-1}\sigma_i^{-1}(w^\top x_i-m_i)^2$.

Note that the whole  posterior approximation is the $k$-dimensional, but the site approximation is the one-dimensional, therefore the degree of freedom are different from the total approximation and the site approximation to make $t$ consistent. The relation between $v$, $\tilde{v}$, and $t$ is given as
\begin{eqnarray}
\frac{1}{t-1}=\frac{v+k}{2}=\frac{\tilde{v}+1}{2}.
\end{eqnarray}

We denote the $\Psi$ and $K$ which is related to site $i$ as $\Psi_i$ and $K_i$. Since $\sigma_i$ is scalar, $K_i=(\tilde{v}\sigma_i)^{-1}$. When we denote $\Psi=(\alpha/|\Sigma|^{1/2})^{1-t}$, then $\Psi_i=(\alpha_i/\sigma_i^{1/2})^{1-t}$.
We denote $\Psi$ and $K$ of whole approximation as $\Psi_w$ and $K_w$.

Let us consider the update of site $j$.
The first step is calculation of cavity distribution, which can be done by
\begin{eqnarray}
\Psi^{\setminus j} K ^{\setminus j}&=&\Psi_w(vV_w)^{-1}-\Psi_j(\tilde{v}\sigma_i)^{-1}x_jx_j^\top,\\
\Psi^{\setminus j} K ^{\setminus j}m^{\setminus j}&=&\Psi_w(vV_w)^{-1}m_w-\Psi_j(\tilde{v}\sigma_i)^{-1}m_jx_j.
\end{eqnarray}
Next step is moment matching. This is calculated in the same way as the ADF update rules. To use the ADF update rule, we have to convert  $\Psi^{\setminus j} K ^{\setminus j}$ and $\Psi^{\setminus j} K ^{\setminus j}m^{\setminus j}$ to $V^{\setminus j}$ and $m^{\setminus j}$, which are covariance matrix and mean of cavity distribution.
When calculating $V^{\setminus j}$ from $\Psi^{\setminus j} K ^{\setminus j}$, we have to be careful that $\Psi^{\setminus j}$ contains the determinant of $V^{\setminus j}$. From the definition,
\begin{eqnarray}
\Psi^{\setminus j} K ^{\setminus j}=\Bigl(\frac{\alpha_j}{|V^{\setminus j}|^{1/2}}\Bigr)^{1-t}(vV^{\setminus j})^{-1}.
\end{eqnarray}
Since $\alpha_j$ and $v$ is the constant, when we put $\frac{{V^{\setminus j}}^{-1}}{|V^{\setminus j}|^{(1-t)/2}}=B$, following relation holds,
\begin{eqnarray}
|V^{\setminus j}|=\Bigl(|B|^{\frac{1}{k}}\Bigr)^{\frac{1}{\frac{t-1}{2}-\frac{1}{k}}}.
\end{eqnarray}
Using this relation, we get $V^{\setminus j}$ and $m^{\setminus j}$.

After moment matching, we get $V_{\mathrm{new}}$ and $m_{\mathrm{new}}$. Next step is the exclusion step of site other than $j$. This step is calculated in the same way as the step of cavity distribution.
\begin{eqnarray}
\Psi_j K_j&=&\Psi_{\mathrm{new}}K_{\mathrm{new}}-\Psi^{\setminus j} K ^{\setminus j},\\
\Psi_j K_j \tilde{m_j}&=&\Psi_{\mathrm{new}}K_{\mathrm{new}}m_{\mathrm{new}}-\Psi^{\setminus j} K ^{\setminus j}m^{\setminus j}.
\end{eqnarray}
To update site parameters, we have to convert $\Psi_j K_j$ and $\Psi_j K_j \tilde{m_j}$ into $\sigma_j$ and $m_j$, which are scalar values. This can be done easily by using the fact that $K_j$ is proportional to $\sigma_j^{-1}x_jx_j^\top$. 

These steps are the update rules for the site approximation. We have to iterate these steps until site parameters converge.

\section{Expectation Propagation for Student-t Process Classification}\label{sec:t-process}
In this section, we show the details of the derivation of EP for the Student-t process classification. The derivation procedure is similar to that of the Gaussian process [2][11][12][13][14].
\subsection{Deriving Update Rules for Student-t Process Classification}
In this subsection, we show the detailed derivation of the update rules for the Student-t process classification.
We denote the prior as $p(f|X)$. In the case of Gaussian process, the prior distribution is a Gaussian distribution whose covariance is specified by the kernel function. In this case, the prior distribution is a Student-t distribution which is specified by the covariance kernel $k(x,x)$ and the degree of freedom $v$. The posterior distribution is given by $p(f|X,y)=\frac{1}{Z}p(f|X)\prod_i p(y_i|f_i)$, where the marginal likelihood is given as $Z=p(y|X)=\int p(f|X)\prod_i P(y_i|f_i)df$ for the i.i.d. situation. In this paper, we consider a binary classification, therefore we use
\begin{eqnarray}
p(y_i|f_i)=l_i (f_i)=\epsilon + (1-2\epsilon)\Theta(y_if_i).
\end{eqnarray}
This is actually the same as BPM, where the input to step function is given as a linear model. In the Student-t process, the input is given as the nonlinear probabilistic process. In this setting, the posterior is intractable; therefore, we have to approximate it.

Following the EP framework, we approximate the posterior by factorizing the posterior in terms of data. To do so, we denote the factorizing term that corresponds to data $i$ as follows.
\begin{eqnarray}
\tilde{l}_i(f_i|\tilde{C}_i,\tilde{\mu}_i,\tilde{\sigma}^2 _i):=\tilde{C}_i\otimes \mathrm{St}(f_i;\tilde{\mu}_i, \tilde{\sigma}^2_i,\tilde{v})
\end{eqnarray}
For simplicity, we denote the unnormalized Student-t like function as $\mathrm{St}(f_i;\tilde{\mu}_i, \tilde{\sigma}^2_i,\tilde{v})$. This is equivalent to $\exp_t(\langle\Phi(f_i),\theta\rangle)$, where $\langle\Phi(f_i),\theta\rangle=\frac{\Psi _i}{1-t}(f_i^\top K_if_i-2\tilde{\mu}_i ^\top K_i f_i)=\frac{\Psi _i}{1-t}(f_i^\top (v\tilde{\sigma}_i)^{-1}f_i-2\tilde{\mu}_i ^\top (v\tilde{\sigma}_i)^{-1}f_i)$. These data corresponding factorizing terms are one-dimensional. 
Note that the whole posterior approximation is the $k$-dimensional, but site approximation is the one dimensional, the same relation as in the BPM between $v$, $\tilde{v}$, and $t$ holds as $\frac{1}{t-1}=\frac{v+k}{2}=\frac{\tilde{v}+1}{2}$.

The q products of this data corresponding term can be expressed as follows:
\begin{eqnarray}
\prod_i \otimes _t \tilde{l}_i(f_i) = \mathrm{St}(\tilde{\mu}, \tilde{\Sigma},v)\otimes _t\prod _i \otimes _t \tilde{C}_i
\end{eqnarray}
Here, we used the property that q products of Student-t distribution become a Student-t distribution. 
In the above expression, $\tilde{\mu}$ is the vector of $\tilde{\mu}_i$ and $\tilde{\Sigma}$ is the diagonal and following relations are given,
\begin{eqnarray}
\tilde{K}^{-1}&=&(v\tilde{\Sigma}),\\
\tilde{\Psi}\tilde{K}&=&\rm{diag}(\Psi_1K_1\dots \Psi_nK_n),\\
&\ &{\rm where\ \ }\tilde{\Psi}=\Biggl(\frac{\Gamma((v+k)/2)}{(\pi v)^{k/2}\Gamma(v/2)|\tilde{\Sigma}|^{1/2}}.\Biggr)^{1-t}.
\end{eqnarray}
Therefore, the total form of the approximation of the posterior can be expressed as follows.
\begin{eqnarray}
q(f|X,y) =\mathrm{St}(\mu, \Sigma,v)\propto p(f|X)\otimes _t \Bigl( \prod_i \otimes _t \tilde{l}_i(f_i)\Bigr)
\end{eqnarray}
From this following relations are obtained, 
\begin{eqnarray}
\Psi K=\Psi_0 K_0+\tilde{\Psi}\tilde{K},\\
\Psi K \mu=\tilde{\Psi}\tilde{K}\tilde \mu.
\end{eqnarray}

We consider the case that we update site $i$.
For implementation, natural parameter based update rule is preferable. Therefore we define the parameter as follows,
\begin{eqnarray}
\tilde{\tau}_i=\tilde{\Psi}_i \tilde{K}_i,
\end{eqnarray}
which is the (i,i) element of $\tilde{\Psi}\tilde{K}$. We also define,
\begin{eqnarray}
\tilde{\nu}_i=\tilde{\Psi}_i \tilde{K}_i\tilde{\mu}_i.
\end{eqnarray}
For the cavity distribution, we define in the same way as,
\begin{eqnarray}
\tau_{-i}&=&\Psi_{-i}\sigma_{-i}^{-2}\tilde{v}^{-1},\\
\nu_{-i}&=&\tau_{-i}\mu_{-i}.
\end{eqnarray}

The first step is to calculate the cavity distribution, we eliminate the effect of site $i$. To do so, we first integrate out non $i$ terms by using the following formula. Let X and Y are random variable that obey the Student-t distribution,
\begin{eqnarray}
\left(
    \begin{array}{r}
      X \\
      Y
    \end{array}
  \right)
\sim \mathrm{St}\Biggl(
\left(
    \begin{array}{r}
      \mu_x \\
      \mu_y
    \end{array}
  \right)
,\left(
    \begin{array}{rr}
      \Sigma _{xx}& \Sigma _{xy} \\
      \Sigma _{yx}& \Sigma _{yy}
    \end{array}
  \right)
,v
\Biggr).
\end{eqnarray}
The marginal distribution X is given as,
\begin{eqnarray}
X\sim \mathrm{St}\bigl(\mu_x, \Sigma _{xx},v\bigr)
\end{eqnarray}
By utilizing the above formula, we get
\begin{eqnarray}
q_{-i}(f_i)&\propto&\int p(f|X)\otimes _t\prod _{j\neq i}\otimes _t l_j(f_j)df_j\\
&\propto&\mathrm{St}(\mu_i,\sigma^2_i,v).
\end{eqnarray}
where, $\mu_i$ is the $i$th element of $\mu$ and $\sigma^2_i$ is the $(i,i)$ element of $\Sigma$. In the above expression, the degree of freedom is $v$ in both the joint distribution and marginal distribution. This is unfavorable for our Student-t process. To make the EP procedure consistent with $t$, we approximate as $q_{-i}(f_i)\propto \mathrm{St}(\mu_i,\sigma'^2_i,\tilde{v})$, $\sigma'^2_i=\sigma^2_i v/\tilde{v}$. Since for a one-dimensional Student-t distribution, its variance is given by $(v\sigma^2_i)^{-1}$, and in this case, $\tilde{v}>v$, approximation by $\sigma'^2_i=\sigma^2_i $ would result in the underestimate of the variance.

We calculate the cavity distribution in the following way,
\begin{eqnarray}
\tau_{-i}&=&\tilde{v}^{-1}{\sigma'}_i^{-2}\Psi_i-\tilde{\tau}_i,\\
\nu_{-i}&=&\tilde{v}^{-1}{\sigma'}_i^{-2}\Psi_i\mu_i-\tilde{\nu}_i.
\end{eqnarray}
Next step is the inclusion of data $i$ to the approximate posterior. This can be done in the same way of BPM. 
To derive the update rule, we have to convert $\tau_{-i}$ and $\nu_{-i}$ into $\sigma^2 _{-i}$ and $\mu_{-i}$.
In this case, the site approximations are one-dimensional, following relation holds,
\begin{eqnarray}
\hat{\mu}_i&=&\mu_{-i}+\sigma^2 _{-i}\alpha,\\
\hat{\sigma_i^2}&=&\sigma^2 _{-i}(r-\alpha \hat{\mu}_i),\\
&\ &\ \ \ \mathrm{where}\ \ \alpha=\frac{\bigl((1-\epsilon)^t-\epsilon^t\bigr)\mathrm{St}(z:,0,1,\tilde{v})}{Z_2\sqrt{\sigma^2 _{-i}}}\ \mathrm{and}\ \ z=\frac{y_i\mu_{-i}}{\sqrt{\sigma^2 _{-i}}},
\end{eqnarray}
where the definition of $Z_2$ and $r$ is same as that of BPM. By using  $\sigma^2 _{-i}$ and $\mu_{-i}$, we can include the data $i$ information.

After the data inclusion step, we exclude the effect other than data $i$. The calculation of this step can be done in the same way as that of cavity distribution,
\begin{eqnarray}
\tilde{\tau}_{i} ^{\mathrm{new}}&=&\tilde{v}^{-1}\hat{\sigma}_i^{-2}\hat{\Psi}_i-\tilde{\tau}_{-i},\\
\tilde{\nu}_{i} ^{\mathrm{new}}&=&\tilde{v}^{-1}\hat{\sigma}_i^{-2}\hat{\Psi}_i\hat{\mu}_i-\tilde{\nu}_{-i}.
\end{eqnarray}

From this $\tilde{\tau}_{i} ^{\mathrm{new}}$, we can update $\tilde{\Psi}\tilde{K}$. Since $\tilde{\Psi}\tilde{K}$ is the diagonal matrix, we just update $(i,i)$ element of  $\tilde{\Psi}\tilde{K}$.

As a final step, we have to update $\Sigma$. To circumvent the calculation of inverse matrix, we put
\begin{eqnarray}
\Delta \tau=-\tilde{\tau}_{i} ^{\mathrm{new}}-\tilde{\tau}_{-i}+\tilde{v}^{-1}\hat{\sigma}_i^{-2}\hat{\Psi}_i
\end{eqnarray}
From this, update of $\Psi K$ is given as,
\begin{eqnarray}
\Psi^{\mathrm{new}} K^{\mathrm{new}}=\Psi^{\mathrm{old}} K^{\mathrm{old}}+\Delta \tau e_ie_i^\top
\end{eqnarray}
where $K^{\mathrm{new}}=(v\Sigma^{\mathrm{new}})^{-1}$ and $K^{\mathrm{old}}=(v\Sigma^{\mathrm{old}})^{-1}$. Here, $\Sigma^{\mathrm{new}}$ is the after the update of $\Sigma$ and $\Sigma^{\mathrm{old}}$ is the before the update of $\Sigma$ and $e_i$ is the unit vector of $i$ th direction. By using the matrix formula, that is, for matrix $A$ and $B$, $(A^{-1}+B^{-1})^{-1}=A-A(A+B)^{-1}A$, we can get the following expression, 
\begin{eqnarray}
{\Psi^{-1}}^{\mathrm{new}}v\Sigma^{\mathrm{new}}={\Psi^{-1}}^{\mathrm{old}}v\Sigma^{\mathrm{old}}-\frac{\Delta \tau}{1+\Delta \tau {\Psi^{-1}}^{\mathrm{old}}v\Sigma^{\mathrm{old}}}s_is_i^\top,
\end{eqnarray}
where $s_i$ is the $i$'s column of ${\Psi^{-1}}^{\mathrm{old}}v\Sigma^{\mathrm{old}}$. From ${\Psi^{-1}}^{\mathrm{new}}v\Sigma^{\mathrm{new}}$, we can get $\Sigma^{\mathrm{new}}$.

These are the update rule of site $i$. We iterate these steps until parameters converge.

\subsection{Hyperparameter Learning}
In this subsection, we refer how to derive hyperparameters, such as the wave-length of covariance functions.

In the usual exponential family and Gaussian process, the hyperparameters can be derived by gradient descent  for the marginal log likelihood after the EP updates end. Following the discussion in [13], we can derive almost the same expression for the gradient of $\log_t Z_{\rm{EP}}^{\frac{2}{3-t}}$. When we consider the gradient of hyperparameter $\psi_i$,
\begin{eqnarray}
\frac{\partial \log _t Z_{\rm{EP}}^{\frac{2}{3-t}}}{\partial \psi_j}=\eta^\top \frac{\partial \theta_{prior}}{\partial \psi_j}-\eta^\top _{prior} \frac{\partial \theta_{prior}}{\partial \psi_j}+\sum_i \frac{\partial \log_t \tilde{C}_i}{\partial \psi_j}
\end{eqnarray}
where, $\theta_{\rm{prior}}$ is the natural parameters of prior distribution and $\eta_{\rm{prior}}$ is the expected sufficient statistics of the prior distribution.
\subsection{Prediction Rule}
In this subsection, we refer to the method of deriving the prediction for the Student-t process classification. After the EP updates end, we have the analytic expression of the approximate posterior distribution as $q(f|X,y) =\mathrm{St}(\mu, \Sigma,v)$.

When a new point $x^*$ is given, we would like to predict its label $y^*$.
First we calculate the latent variable $f^*$ of $x^*$.
To get the expression of $f^*$, we use the following lemma in [4],
\begin{lmm}
If $X \sim \mathrm{St}(\mu,\Sigma,v)$, and $x_1\in R^{n_1}$, $x_2\in R^{n_2}$ express the first $n_1$ and remaining $n_2$ entries of X respectively. Then
\begin{eqnarray}
x_2|x_1\sim \mathrm{St}\Bigl(\tilde{\mu}_2,\frac{v+\beta_1}{v+n_1}\times \tilde{\Sigma}_{22},v+n_1\Bigr),
\end{eqnarray}
where $\tilde{\mu}_2=\Sigma_{21}\Sigma_{11}^{-1}(x_1-\mu_1)+\mu_1$, $\tilde{\Sigma}_{22}=\Sigma_{22}-\Sigma_{21}\Sigma_{11}^{-1}\Sigma_{12}$, $\beta_1=(x_1-\mu_1)^\top K_{11}^{-1}(x_1-\mu_1)$.
\end{lmm}
We consider the following expression,
\begin{eqnarray}
p(\tilde{f}|X,x^*)=\int p(\tilde{f}|f,x^*)p(f|X)df.
\end{eqnarray}
The mean of $p(\tilde{f}|X,x^*)$ is given by
\begin{eqnarray}
\mathrm{E}[\tilde{f}]&=&\int \mathrm{E}[p(\tilde{f}|f,x^*)]p(f|X)df\\
&=&\int k^\top\Sigma^{-1}fp(f|X)df\\
&=&k^\top\Sigma^{-1}\mu
\end{eqnarray}
where, $k=[k(x^*,x_1),\dots k(x^*,x_n)]^\top$. Therefore strict classification of $x^*$ is given by
\begin{eqnarray}
\mathrm{sign}\bigl(\mathrm{E}[\tilde{f}]\bigr)=\mathrm{sign}\bigl(k^\top\Sigma^{-1}\mu)
\end{eqnarray}
Using this expression, we get the decision boundary.
\end{document}